\documentclass[10pt,twocolumn,letterpaper]{article}

\usepackage[pagenumbers]{cvpr} 

\definecolor{cvprblue}{rgb}{0.21,0.49,0.74}
\definecolor{kevin}{rgb}{0.74,0.49,0.21}

\usepackage[pagebackref,breaklinks,colorlinks,allcolors=cvprblue]{hyperref}
\usepackage{url}
\usepackage{graphicx}
\usepackage{amsmath}
\usepackage{amssymb}
\usepackage{booktabs}
\usepackage{color}
\usepackage{colortbl}
\usepackage{xcolor}
\usepackage{bbm}
\usepackage{bm}
\usepackage{graphicx}
\usepackage{multirow}
\usepackage{float}
\usepackage{makecell}
\usepackage[title]{appendix}
\usepackage{makecell}    
\usepackage{pifont}

\def\paperID{} 
\def\confName{CVPR}
\def\confYear{2026}

\def\ours{Re-Align}
\def\bagel{BAGEL}
\def\omnigen2{OmniGen2}

\definecolor{myblue}{RGB}{210, 225, 255}

\title{Re-Align: Structured Reasoning-guided Alignment \\ for In-Context Image Generation and Editing}

\small
\vspace{-5pt}
\author{
Runze He$^{1,2,3}$,
Yiji Cheng$^{1}$,
Tiankai Hang$^{1}$,
Zhimin Li$^{1}$,
Yu Xu$^{1}$, 
Zijin Yin$^{1}$, 
Shiyi Zhang$^{1}$, \\
Wenxun Dai$^{1}$,
Penghui Du$^{3}$,
Ao Ma$^{3}$,
Chunyu Wang$^{1}$\footnotemark[2],
Qinglin Lu$^{1}$,
Jizhong Han$^{2,3}$,
Jiao Dai$^{2,3}$\footnotemark[3] \\
\vspace{5pt}
$^{1}$Hunyuan, Tencent \quad
$^{2}$IIE, CAS \quad
$^{3}$UCAS \\
\vspace{5pt}
Project Page: \url{https://hrz2000.github.io/realign}
}

\begin{document}
\twocolumn[{
\renewcommand\twocolumn[1][]{#1}
\maketitle

\begin{center}
    \centering
    \vspace*{-.8cm}
    \includegraphics[width=.99\textwidth]{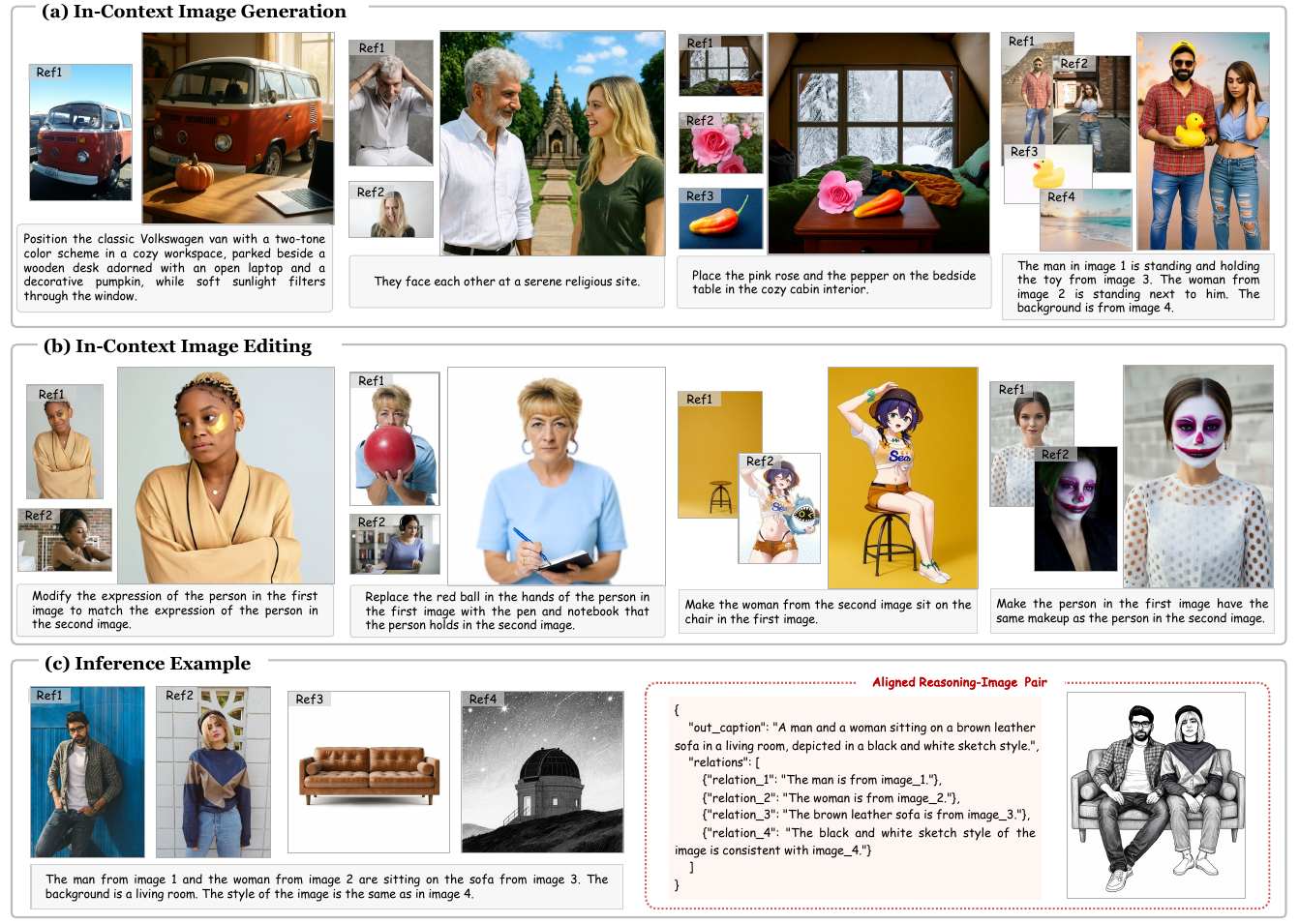}
    \vspace*{-.2cm}
    \captionof{figure}{Our proposed \ours{} supports image synthesis conditioned on flexible image-text interleaved prompts, namely \textbf{a)} in-context image generation, also referred to as subject-driven image generation, and \textbf{b)} in-context image editing, also referred to as reference-based image editing.
    \textbf{c)} An inference example from \ours{}, including an aligned reasoning–image pair. 
    The reasoning text is converted from XML to JSON for clearer visualization.
    }
\label{fig:fig1}
\vspace{-8pt}
\end{center}

}]

\footnotetext{
\footnotemark[2]{Project lead.}\quad
\footnotemark[3]{Corresponding author.}
}

\begin{abstract}

In-context image generation and editing (ICGE) enables users to specify visual concepts through interleaved image-text prompts, demanding precise understanding and faithful execution of user intent.
Although recent unified multimodal models exhibit promising understanding capabilities, these strengths often fail to transfer effectively to image generation.
We introduce {Re-Align}, a unified framework that bridges the gap between understanding and generation through structured reasoning-guided alignment.
At its core lies the {In-Context Chain-of-Thought (IC-CoT)}, a structured reasoning paradigm that decouples {semantic guidance} and {reference association}, providing clear textual target and mitigating confusion among reference images.
Furthermore, Re-Align introduces an effective RL training scheme that leverages a surrogate reward to measure the alignment between structured reasoning text and the generated image, thereby improving the model’s overall performance on ICGE tasks.
Extensive experiments verify that \ours{} outperforms competitive methods of comparable model scale and resources on both in-context image generation and editing tasks.  

\end{abstract}
\section{Introduction}
\label{sec:intro}

In recent years, the field of image synthesis~\cite{rombach2022stablediffusion,flux,wu2025qwenimagetechnicalreport,gal2022stylegan,VAR,chang2022maskgit,sun2024autoregressive_llamagen,ho2020ddpm,song2021ddim,liu2022flowmatching} has attracted widespread attention from the research community. Among them, diffusion models~\cite{ho2020ddpm,song2021ddim,liu2022flowmatching} have made significant progress due to their ability to generate diverse and high-quality samples.
Given that pure text prompts often fail to accurately express visual concepts defined by reference images, image-conditioned visual generation~\cite{ruiz2022dreambooth,kumari2022customdiffusion,gal2022ti,uno,tao2025instantcharacter} has also been extensively explored. Recently, with the ability to process interleaved image–text inputs, in-context image generation and editing (ICGE) has become increasingly popular.

\begin{figure}[!t]
    \centering
    \includegraphics[width=\linewidth]{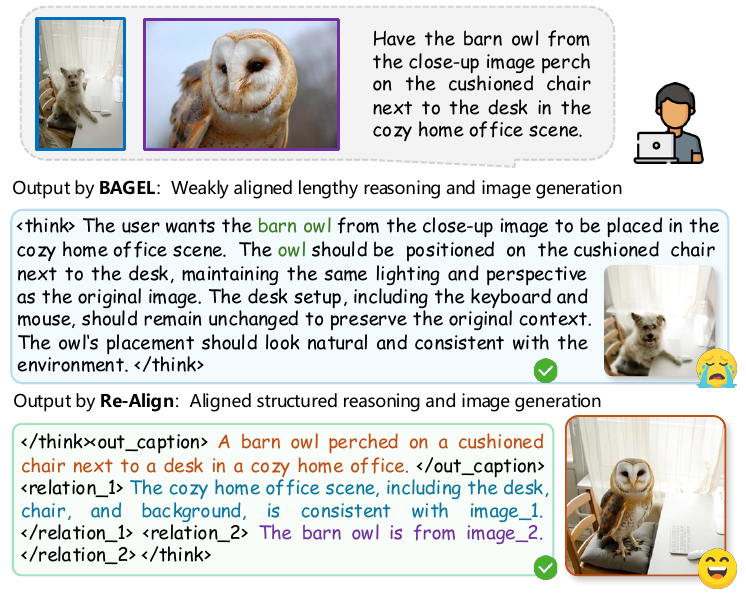}
    \caption{Comparison of the reasoning paradigms of \bagel{} and \ours{}. While \bagel{} exhibits competent reasoning abilities, the resulting images fail to reflect its reasoning process in the complex image-text interleaved prompt. In contrast, \ours{} achieves strong reasoning–generation alignment, facilitated by the structured IC-CoT. }
    \label{fig:cot}
    \vspace{-15pt}
\end{figure}

Nevertheless, implementing ICGE is non-trivial, as it requires both \textbf{precise understanding} of the complex interleaved inputs and \textbf{faithful execution} of the user's intent. Reasoning mechanisms that are effective for text-to-image and image editing, however, fail to function effectively in ICGE tasks.
For example, the leading native multimodal model \bagel{}~\cite{deng2025bagel} can accurately interpret instructions and produce plausible reasoning, yet the final generated image fails to align with this reasoning, as shown in Figure~\ref{fig:cot}. 
This suggests that although the reasoning ability is strong, it has not yet helped downstream image generation, and there is a misalignment between the two.

Building on these insights, we propose \ours{}, a unified framework designed for in-context image generation and editing with structured \textbf{Re}asoning-guided \textbf{Align}ment.
\ours{} adopts a structured reasoning mechanism, namely In-Context Chain-of-Thought (IC-CoT), which explicitly decomposes the reasoning process into semantic guidance and reference association and is uniformly applied to both image generation and editing.
The former provides a clear textual target for image generation, partly simplifying the image-text interleaved task into a text-to-image generation; the latter analyzes the role of each reference image within the multi-image context to prevent reference confusion.
To further enhance model's performance on complex interleaved prompts, we employ Group Relative Policy Optimization (GRPO) with a surrogate reward that measures the correspondence between the CoT context and the resulting image.
The reasoning-induced diversity strategy is proposed to improve the diversity of samples between groups, thereby stabilizing the training of GRPO.
To support model training, we develop an automated data construction and filtering pipeline, yielding \ours{}-410K, a high-quality ICGE dataset with IC-CoT annotations spanning multiple in-context image generation and editing tasks.

Our contributions are summarized as follows:
(1) We present \ours{}, which achieves state-of-the-art performance among methods with comparable resources and model scales on both in-context image generation and editing tasks.
(2) We propose a structured reasoning paradigm, IC-CoT, which provides a clear target for visual generation through decoupled semantic guidance and reference association.
(3) We further introduce a surrogate reward that measures the alignment between the reasoning context and the generated image, along with a reasoning-induced diversity strategy, enabling effective policy optimization for improved model performance.

\section{Related Works}

\subsection{In-Context Image Generation and Editing}
Instead of pursuing isolated, single-purpose conditional generation, such as {image customization}~\cite{ruiz2022dreambooth,gal2022ti,kumari2022customdiffusion,li2023blipdf} or {image editing}~\cite{brooks2023instructpix2pix,Isola2017pix2pix,Zhang2023MagicBrush,he2024freeedit,zhao2024ultraedit}, {in-context image generation and editing} focuses on general-purpose generation tasks guided by flexible interleaved image-text prompts.
Closed-source systems such as GPT-4o~\cite{openai2025gptimage} and Nano Banana~\cite{google2025nanobanana} have exhibited remarkable performance on such tasks.
Meanwhile, open-source models~\cite{wu2025omnigen2,cui2025emu35,deng2025bagel,ye2025echo,xia2025dreamomni2} are steadily progressing toward this goal. 
{\bagel{}}~\cite{deng2025bagel}, as a native multimodal foundation model, inherently supports simple in-context image generation and editing tasks. 
{OmniGen2}~\cite{wu2025omnigen2}, conditioned on the hidden states of an MLLM~\cite{bai2025qwen25vl}, demonstrates versatile image generation capabilities. 
Our concurrent work, {DreamOmni2}~\cite{xia2025dreamomni2}, employs a joint training framework for the generation/editing model and MLLM, sharing the same goal as ICGE.
Despite their promising results, these methods remain inadequate when handling complex image–text interleaved instructions.

\subsection{Unified Understanding and Generation}
Recently, unified models~\cite{wang2024emu3,wu2024janus,chen2025janus_pro,ma2024janusflow,deng2025bagel,wu2025omnigen2,zhou2024transfusion,chen2025blip3o,pan2025metaquery,xie2025showo2,he2025plangen,cui2025emu35} that integrate both understanding and generation capabilities have been extensively explored. 
Among them, Emu3~\cite{wang2024emu3}, Janus~\cite{wu2024janus}, and Janus-Pro~\cite{chen2025janus_pro} model understanding and generation solely through next-token prediction. 
Show-o~\cite{xie2024showo} unifies autoregressive and discrete diffusion modeling, enabling adaptive handling of inputs and outputs across mixed modalities. 
Several approaches~\cite{wu2025omnigen2, pan2025metaquery, chen2025blip3o} employ frozen LLMs for understanding and an additional DiT~\cite{peebles2023dit} for image generation, thereby reducing training overhead and mitigating interference between the two capabilities. 
Transfusion~\cite{zhou2024transfusion} and \bagel{}~\cite{deng2025bagel} employ autoregressive modeling for understanding and diffusion modeling for image generation within a single transformer architecture, with \bagel{} further introducing a Mixture-of-Transformers structure to enhance performance.

\subsection{Reinforcement Learning for Visual Generation}
Reinforcement learning (RL) has recently achieved remarkable progress in the development of large language models (LLMs)~\cite{Brown2020GPT3, dubey2024llama3.1-paper, hui2024qwen2.5-paper, shao2024deepseekmath, guo2025deepseekr1}, which has in turn spurred growing interest in applying RL techniques to visual generation tasks~\cite{fan2023dpok, ddpo, xu2023imagereward, wallace2024dpo, liu2025flowgrpo, xue2025dancegrpo, shen2025srpo}.  
Recent research has particularly explored the use of {Group Relative Policy Optimization (GRPO)}~\cite{shao2024deepseekmath}, owing to its ability to eliminate the need for a separate value network, thereby improving memory efficiency compared with {Proximal Policy Optimization (PPO)}~\cite{schulman2017ppo}.  
Building on this, {FlowGRPO}~\cite{liu2025flowgrpo} and {DanceGRPO}~\cite{xue2025dancegrpo} extend the GRPO to image and video generation.
However, existing RL-based approaches predominantly focus on optimizing text-conditioned generation, and still lack effective reward design and comprehensive experimental validation for more {complex in-context image generation and editing} tasks.
\section{Method}

\begin{figure*}[!t]
    \centering
    \includegraphics[width=\linewidth]{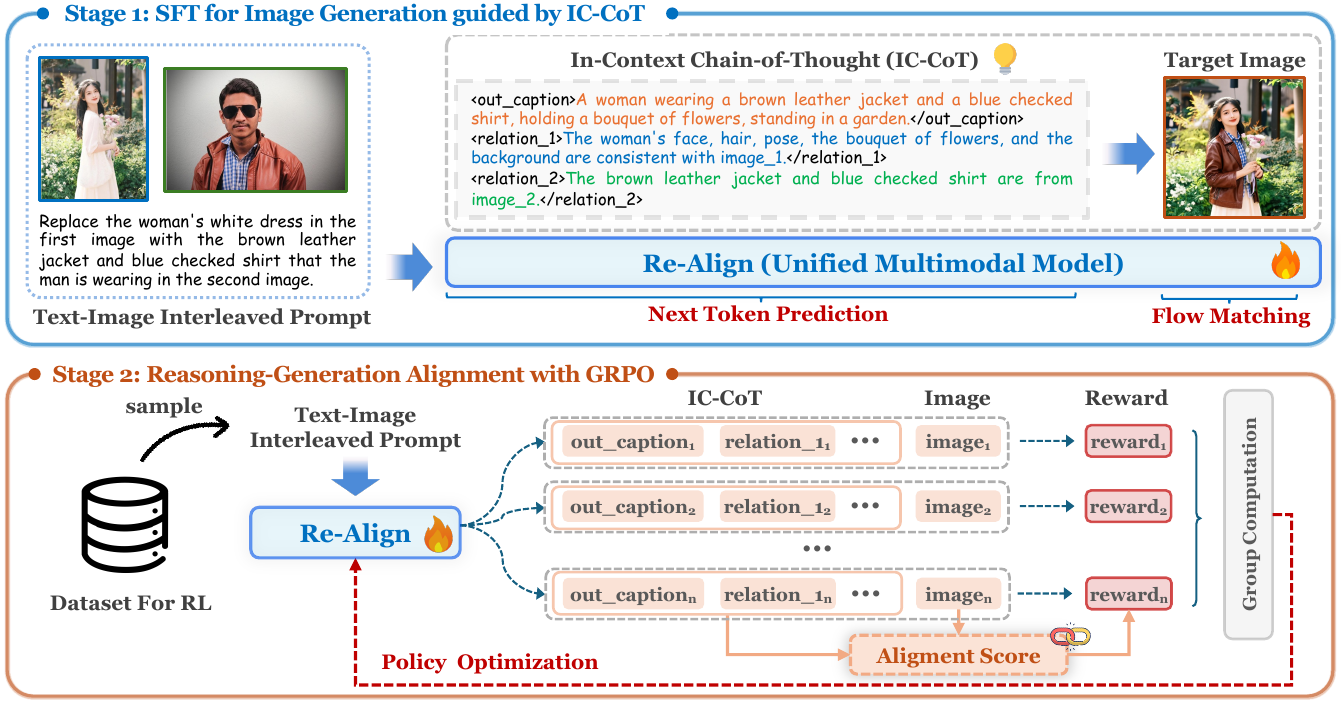}
    \vspace{-6mm}
    \caption{
    The two-stage training pipeline of \ours{}. First, we perform supervised fine-tuning on carefully curated training data to enable the model to generate images guided by IC-CoT reasoning. Next, we apply policy optimization to further enhance reasoning–generation consistency, using an alignment score between the structured IC-CoT and the corresponding generated image.
    }
    \label{fig:pipeline}
    \vspace{-10pt}
\end{figure*}

\subsection{Overview}
As illustrated in Figure~\ref{fig:pipeline}, \ours{} serves as a unified framework designed for in-context image generation and editing, based on the architecture of the multimodal foundation model \bagel{}~\cite{deng2025bagel}.
Given a image-text interleaved prompt $\bm{P}$, including serveral reference images, a text instruction which couples visual concepts like ``\textit{Replace the hat in the first image with the cup in the second image}", \ours{} generates structured reasoning text, i.e. In-Context Chain-of-Thought, denoted as $\bm{R}=\{\bm{r}_1,\bm{r}_2,...,\bm{r}_M\}$ with $M$ reasoning tokens, and resulting image $\bm{I}$ sequentially.

Specifically, we maximize the likelihood of reasoning tokens given prompt $\bm{P}$ and all previously generated reasoning tokens by employing the standard language modeling objective:
\begin{equation}
\vspace{-5pt}
\mathcal{L}_{\text{cot}}(\theta)=\sum_i\log p_\theta(\bm{r}_i|\bm{P},\bm{r}_{<i}),
\vspace{-3pt}
\end{equation}
where $p$ indicates the conditional probability of the model, parameterized by weights $\theta$.

Let $\boldsymbol{x}_0 \sim p_0$ be a sample from the real data distribution and $\boldsymbol{x}_1 \sim p_1$ a noise sample from the Gaussian distribution.
We adopt the Rectified Flow~\cite{liu2022flowmatching} to learn the image generation following \bagel{}~\cite{deng2025bagel}, with the objective:
\begin{equation}\mathcal{L}_{\text{img}}(\theta)=\mathbb{E}_{t,\boldsymbol{x}_0\boldsymbol{\sim}p_0,\boldsymbol{x}_1\boldsymbol{\sim}p_1}\left[\|\boldsymbol{v}-\boldsymbol{v}_\theta(\boldsymbol{x}_t,t,\bm{P},\bm{R})\|^2\right],\end{equation}
where $\bm{x}_t = (1 - t)\bm{x}_0 + t\bm{x}_1$ for $t \in [0, 1]$ denotes noisy data, $\bm{v}_\theta(\bm{x}_t, t, \cdot)$ is the predicted velocity field, and $\bm{v} = \bm{x}_1 - \bm{x}_0$ is the target velocity field.

\subsection{In-Context Chain-of-Thought}
\label{sec:cot}
Previous works~\cite{deng2025bagel,fang2025got,wang2025mint} have demonstrated the benefits of introducing the reasoning capability into visual generation.
Nevertheless, these approaches are limited to text-conditioned image generation and editing, while effective reasoning in more complex ICGE tasks remains unexplored. 
When faced with complex interleaved image-text prompts, the leading unified multimodal model \bagel{}~\cite{deng2025bagel} fails to produce consistent reasoning and image outputs, indicating that its reasoning mechanism is not effectively utilized. Thereby, we aim to leverage the reasoning mechanism to bridge the gap between the model’s understanding and generation abilities. Specifically, we propose In-Context Chain-of-Thought (IC-CoT), which is a structured reasoning framework, including two complementary components: \textit{semantic guidance} and \textit{reference association}.
The former provides an explicit caption to facilitate image generation under complex conditions, while the latter captures the associative relationships between each reference image and the target to prevent reference confusion.

\noindent\textbf{Semantic Guidance}
Interleaved image-text prompts often convey complex and implicit user intentions, making direct image generation challenging due to the intricate semantic interactions between visual and textual elements. 
IC-CoT explicitly predicts the caption of the resulting image as part of its reasoning process, starting with \texttt{<out\_caption>} and ending with \texttt{</out\_caption>}. 
The predicted caption provides direct semantic guidance for subsequent image generation, while remaining compatible with both instructional and descriptive user inputs. 
By incorporating the predicted caption into the reasoning process, complex in-context image generation and editing tasks can be partially reduced to text-conditioned generation, thereby easing the learning process.

\noindent\textbf{Reference Association}
ICGE features user-provided reference images. However, the flexible nature of interleaved image-text inputs often leads users to omit explicit references to image indices or corresponding subjects, and instead use ambiguous expressions such as ``\textit{put them together}'', making it more difficult for the model to interpret the user’s intended output.
To address this, IC-CoT introduces reference associations in the reasoning process, starting with \texttt{<relation\_i>} and ending with \texttt{</relation\_i>} for the reference image $\bm{i}$. 
Each reference association specifies the role of the corresponding reference image in generating the final output, and the number of associations matches the number of provided reference images.

The structured IC-CoT plays a key role in bridging the gap between the model’s understanding and generation capabilities.
Compared to the prompt-expansion paradigm such as \bagel{}~\cite{deng2025bagel}, IC-CoT employs a compact structured representation to provide clear semantic and reference cues for image generation, thereby reducing ambiguity and lowering both training and inference overhead.
Moreover, the structured IC-CoT enables effective extraction of key elements, facilitating the subsequent alignment stage.

\begin{figure*}[!t]
    \centering
    \includegraphics[width=\linewidth]{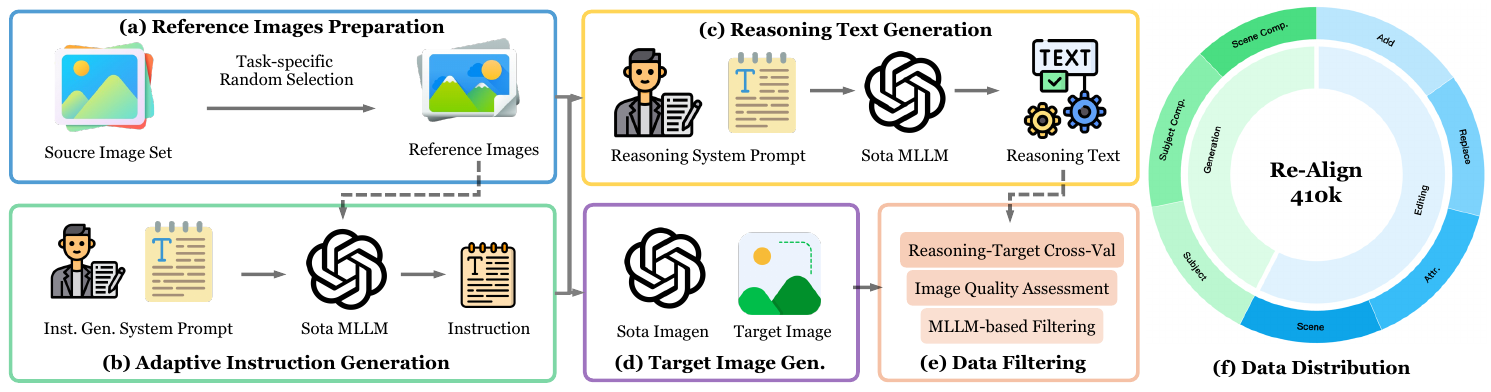}
    \vspace{-6mm}
    \caption{The data construction pipeline of \ours{}-410K and its task distribution. \textbf{a)} reference images preparation,
    \textbf{b)} adaptive instruction generation,
    \textbf{c)} reasoning text generation,
    \textbf{d)} target image generation,
    \textbf{e)} data filtering, and
    \textbf{f)} the data distribution of \ours{}-410K.
    }
    \label{fig:data}
    \vspace{-1mm}
\end{figure*}

\subsection{Reasoning-Generation Alignment}
\label{sec:grpo}

Despite the remarkable progress of GRPO~\cite{shao2024deepseekmath} in visual generation~\cite{xue2025dancegrpo,liu2025flowgrpo}, existing reward models are typically designed for text-conditioned generation. 
In contrast, ICGE tasks involve interleaved image–text inputs, diverse generation and editing tasks, and multidimensional evaluation criteria, making the construction of a dedicated reward model extremely costly and complex. 
Therefore, applying reinforcement learning to ICGE remains challenging.

\noindent\textbf{Surrogate Reward for ICGE}
Instead of designing task-specific reward models, we introduce a surrogate reward that measures the alignment between the reasoning context and the generated image, thus indirectly improving the model’s overall performance.
Aligning unstructured reasoning texts with images is challenging, as it is difficult to extract representative semantic content that is useful for bridging the two modalities. 
Thanks to the IC-CoT’s structured format, we can readily extract the semantic guidance enclosed within \texttt{<out\_caption>} and \texttt{</out\_caption>}, i.e., the predicted caption $\bm{c}$. The image-text similarity between $\bm{c}$ and the generated image $\bm{x}$ then serves as the reward signal $\bm{s}$, which is computed as follows:
\begin{equation}
\bm{s}(\bm{x}, \bm{c}) = \frac{
  \mathcal{E}(\bm{x})^\top \mathcal{T}(\bm{c})
}{
  \|\mathcal{E}(\bm{x})\| \cdot \|\mathcal{T}(\bm{c})\|
},
\end{equation}
Here, $\mathcal{E}$ and $\mathcal{T}$ are the image and text encoders of CLIP~\cite{radford2021clip}, respectively, and $\|\cdot\|$ denotes the L2 norm.



\noindent\textbf{Reasoning-Induced Diversity Strategy}
In ICGE tasks, the explicit visual concepts provided in the input impose strong constraints on the generation process, thereby reducing sample diversity compared with text-conditioned generation. When the differences among generated samples become small, the reward variance also diminishes; after normalization, even minor fluctuations may be disproportionately amplified, ultimately hindering the model’s ability to learn effectively from the reward signal.
Prior works~\cite{xue2025dancegrpo,liu2025flowgrpo} attempt to enlarge sample diversity by increasing the SDE noise scale, but excessive noise often degrades image quality. In contrast, we generate distinct IC-CoT reasoning chains for each sample within a group, introducing diverse reasoning trajectories that naturally diversify the outputs. This strategy increases reward variance in a controlled manner, providing more informative learning signals and thereby stabilizing the training process.

\subsection{Dataset Construction}
\label{sec:data}

As shown in Figure~\ref{fig:data}, to support model training, we introduce \ours{}-410K, a high-quality collection covering the task types summarized in Table~\ref{tab:data}.
The dataset is constructed via an automated data construction pipeline that integrates advanced MLLMs~\cite{gemini2p5,achiam2023gpt4} and state-of-the-art image generation models~\cite{gpt4o}.

\noindent\textbf{Reference Images Preparation}
Unlike conventional single–image conditioned generation or editing tasks~\cite{brooks2023instructpix2pix,Sheynin2023emu_edit,Zhang2023MagicBrush,ye2023ipadapter,zhang2025icedit,zhang2023controlnet,mo2024freecontrol}, ICGE supports flexible interleaving of multiple image and text inputs. This setting demands a dataset with diverse reference-image combinations. To accommodate this requirement, we construct a source image pool covering characters, objects, and scenes, from which multiple references are sampled according to each task type.
For subject-reference tasks, the sampled references are drawn from character and object categories, whereas scene-centric tasks additionally incorporate scene images. For attribute-reference editing, references are selected with greater flexibility to support a broad spectrum of attribute-guided modifications.


\begin{table}[ht]
\centering
\begin{tabular}{@{}p{\linewidth}@{}}
\toprule
\textbf{1. In-Context Image Generation} \\
\begin{minipage}[t]{\linewidth}
\begin{itemize}[leftmargin=*, nosep, after=\strut]
    \item \textbf{Subject-driven Generation}: Generate a referenced subject in a novel context.
    \item \textbf{Subject-Subject Compositional Generation}: Combine multiple referenced subjects within a new scene.
    \item \textbf{Subject-Scene Compositional Generation}: Place multiple referenced subjects into a referenced scene under a new context.
\end{itemize}
\end{minipage} \\
\midrule
\textbf{2. In-Context Image Editing} \\
\begin{minipage}[t]{\linewidth}
\begin{itemize}[leftmargin=*, nosep, after=\strut]
    \item \textbf{Reference Subject Editing}: Add a referenced subject to an input image or replace an existing subject with the referenced one.
    \item \textbf{Reference Attribute Editing}: Transfer attributes from a reference, such as texture, pose, style or other visual characteristics, to modify the appearance of the subject (Local) or target image (Global).
    \item \textbf{Reference Scene Editing}: Modify the scene of an image based on a referenced one.
\end{itemize}
\end{minipage} \\
\bottomrule
\end{tabular}
\caption{Overview of the tasks covered in \ours{}-410K.}
\label{tab:data}
\vspace{-10pt}
\end{table}

\noindent\textbf{Adaptive Instruction Construction}
Next, we generate instructions tailored to each group of reference images. Since fixed manual rules cannot capture the diversity of visual content, we leverage the advanced Gemini 2.5~\cite{gemini2p5} for adaptive instruction generation. A carefully designed system prompt guides the MLLM to produce executable instructions conditioned on the input images, while additionally encouraging attention to secondary visual details to increase the complexity and richness of the generated instructions.

\noindent\textbf{Reasoning Text Generation}
Unlike previous works~\cite{wu2025omnigen2,ye2025echo,xia2025dreamomni2}, which focus solely on constructing input–output pairs while neglecting the underlying reasoning process, we additionally prompt the MLLM to generate the structured IC-CoT introduced in Section~\ref{sec:cot}. The reference images together with the corresponding instructions are provided to the MLLM, which produces the structured reasoning output under a predefined system prompt. Notably, the target image is intentionally omitted during this stage, as introducing additional visual inputs may increase hallucination and impair the model’s ability to correctly interpret complex multi-image relationships.

\noindent\textbf{Target Image Generation}
As demonstrated in prior studies~\cite{chen2025blip3o,chen2025sharegpt,ye2025echo}, data generated by the state-of-the-art image generator GPT-4o~\cite{gpt4o} effectively handles complex and long-tail visual generation scenarios. Accordingly, we feed the reference image group along with the corresponding generated instructions into GPT-4o to synthesize the target images. 
Unlike video frame extraction methods like OmniGen2~\cite{wu2025omnigen2}, which are largely confined to in-context image generation, our approach can effectively handle a wider variety of editing tasks.

\noindent\textbf{Data Filtering}
To ensure high data quality, we adopt a multidimensional filtering strategy. 
First, we leverage the structured IC-CoT format to compute image–text similarity between the caption predicted by the reasoning process and the target image, with low similarity indicating a misalignment between reasoning and generation. 
Second, we assess visual quality using image aesthetics~\cite{LAION-Aesthetics} and human preference metrics~\cite{xu2023imagereward,kirstain2023pick}.
Third, we evaluate instruction following and semantic consistency capability using OmniContextScore~\cite{wu2025omnigen2}. 
Samples below any threshold are discarded, removing approximately 20\% of the data and resulting in a final dataset of 410K high-quality samples.

\section{Experiments}

\begin{figure*}[!t]
    \centering
    \includegraphics[width=\linewidth]{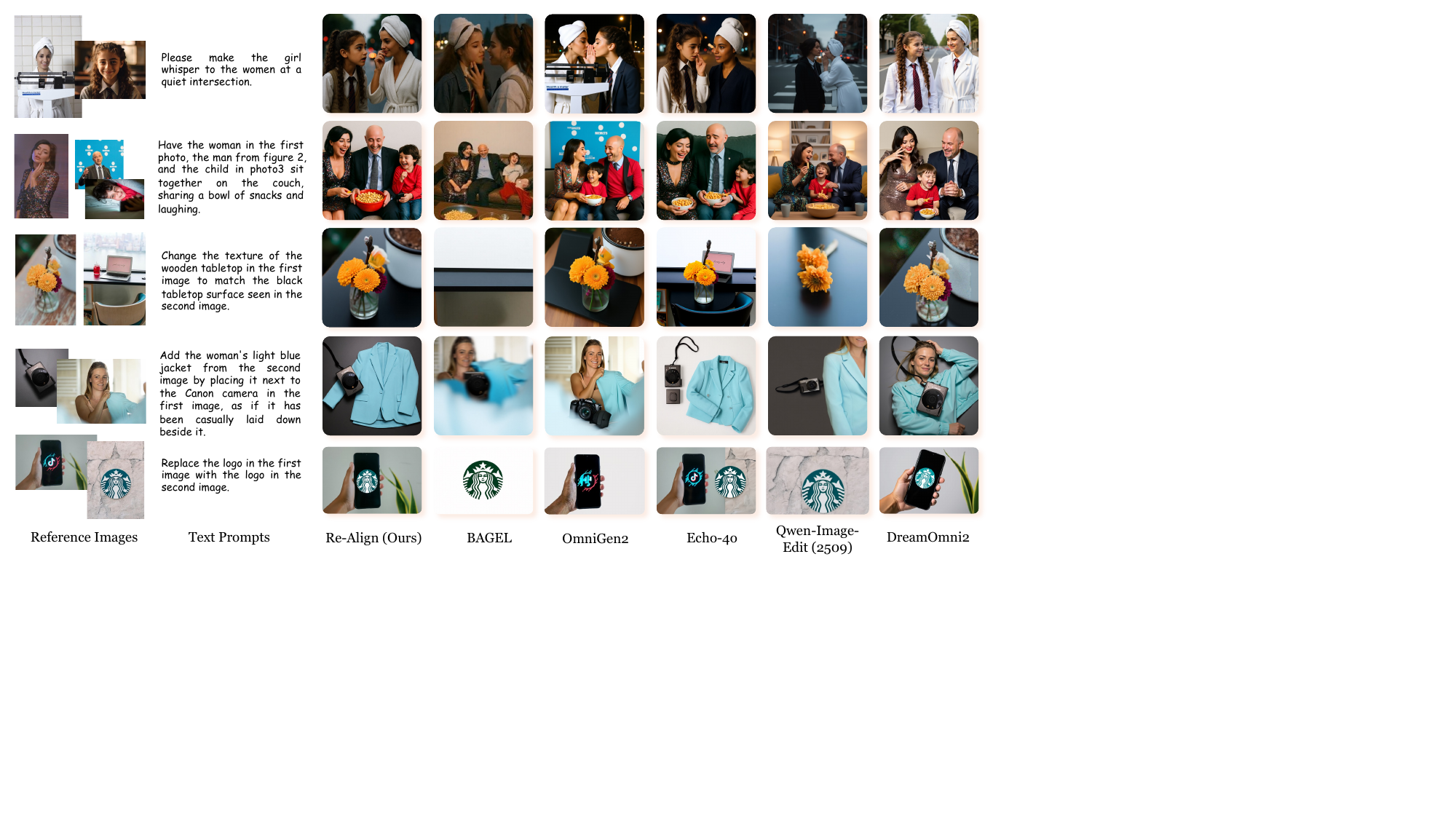}
    \vspace{-6mm}
    \caption{Qualitative comparisons of proposed \ours{} with \bagel{}~\cite{deng2025bagel}, OmniGen2~\cite{wu2025omnigen2}, Echo-4o~\cite{ye2025echo}, Qwen-Image-Edit(2509)~\cite{wu2025qwenimagetechnicalreport} and DreamOmni2~\cite{xia2025dreamomni2} on the in-context image generation and editing tasks.}
    \label{fig:comp}
\end{figure*}

\subsection{Experimental Setup}

\noindent \textbf{Baselines.}
We compare \ours{} with several recent representative methods widely recognized for in-context image generation and editing, including:
(1) \bagel{}~\cite{deng2025bagel}, a foundational model that natively supports multimodal understanding and generation;
(2) OmniGen2~\cite{wu2025omnigen2}, a versatile generative model providing a unified solution for diverse tasks;
(3) Echo-4o~\cite{ye2025echo}, fine-tuning \bagel{} on a high-quality synthetic image dataset;
(4) Qwen-Image-Edit(2509)~\cite{wu2025qwenimagetechnicalreport}, a version of Qwen-Image-Edit supporting multi-image input; and
(5) DreamOmni2~\cite{xia2025dreamomni2}, a concurrent work focusing on multimodal instruction-based image editing and generation.

\noindent \textbf{Implementation Details.}
Proposed \ours{} builds upon \bagel{}~\cite{deng2025bagel} and is compatible with other models~\cite{chen2025janus_pro,zhou2024transfusion,cao2025hunyuanimage3} that provide unified understanding and generation capabilities.
We employ a mixed training strategy, both with and without IC-CoT, to provide flexibility during inference.
The SFT stage is trained for 100{,}000 steps on 64 NVIDIA H20 GPUs with a learning rate of $5\times10^{-6}$, while the reasoning–generation alignment stage is trained for 200 steps with a group size of 32 and a learning rate of $1\times10^{-6}$, which is sufficient to ensure alignment convergence and avoid unnecessary reward hacking in subsequent training.
By default, images are generated at 1024{×}1024 resolution using 50 denoising steps, following~\cite{deng2025bagel}.

\noindent \textbf{Benchmarks.}
We evaluate models' ICGE capabilities on two mainstream benchmarks: OmniContext~\cite{wu2025omnigen2} and DreamOmni2Bench~\cite{xia2025dreamomni2}. 
OmniContext provides a comprehensive suite for evaluating in-context image generation across diverse scenarios. 
In contrast, DreamOmni2Bench offers a large collection of generation and editing tasks, with one to five reference images as input, covering diverse editing settings ranging from local and global attributes to object-level manipulations.

\noindent \textbf{Evaluation Metrics.}
Similar to VIEScore~\cite{ku2023viescore} in image editing, OmniContext~\cite{wu2025omnigen2} uses the multimodal large language model GPT-4.1~\cite{gpt4-1} as an automatic evaluator for in-context visual generation. 
It includes three metrics: Prompt Following (PF), measuring whether the generated image fulfills the editing intent; Subject Consistency (SC), evaluating the consistency of visual concepts between the generated image and reference images; and an Overall Score, computed as the geometric mean of PF and SC. 
Since the official evaluation code for DreamOmni2Bench~\cite{xia2025dreamomni2} is not yet available, we employ OmniContext's metric framework to evaluate model performance on this benchmark as well.

\begin{table*}[t]
    \centering
    \resizebox{0.99\linewidth}{!}{
    \begin{tabular}{l|cc|ccc|ccc|c}
        \toprule
        \multirow{2}{*}{\bf Model} & \multicolumn{2}{c|}{\bf SINGLE} & \multicolumn{3}{c|}{\bf MULTIPLE} & \multicolumn{3}{c|}{\bf SCENE} & \multirow{2}{*}{\bf Average$\uparrow$}\\ 
        \cmidrule(lr){2-9}
        & Character & Object & Character & Object & Char. + Obj. & Character & Object & Char. + Obj. & 
        \\
        \midrule
        FLUX.1 Kontext [Max]~\cite{labs2025flux} & 8.48 & 8.68 & - & - & - & - & - & - & - 
        \\
        Gemini 2.0 Flash~\cite{gemini-2.0-flash} & 5.06 & 5.17 & 2.91 & 2.16 & 3.80 & 3.02 & 3.89 & 2.92 & 3.62
        \\
        Gemini 2.5 Flash Image~\citep{google2025nanobanana} & 8.62 & 8.91 & 7.88 & 8.92 & 7.39 & 7.29 & 7.05 & 6.68 & 7.84
        \\
        GPT-4o~\cite{gpt4o} & \textbf{8.90} & 9.01 & \textbf{9.07} & 8.95 & 8.54 & \textbf{8.90} & 8.44 & \textbf{8.60} & 8.80
        \\
        Emu3.5~\cite{cui2025emu35} & 8.72 & \textbf{9.46} & 8.65 & \textbf{9.09} & \textbf{8.78} & 8.78 & \textbf{8.89} & 8.15 & \textbf{8.82}
        \\
        \midrule
        OmniGen~\cite{xiao2025omnigen} & 7.21 & 5.71 & 5.65 & 5.44 & 4.68 & 3.59 & 4.32 & 5.12 & 4.34
        \\
        InfiniteYou~\cite{jiang2025infiniteyou} & 6.05 & - & - & - & - & - & - & - & - 
        \\
        UNO~\cite{uno} & 6.60 & 6.83 & 2.54 & 6.51 & 4.39 & 2.06 & 4.33 & 4.37 & 4.71
        \\
        BAGEL~\cite{deng2025bagel} & 5.48 & 7.03 & 5.17 & 6.64 & 6.24 & 4.07 & 5.71 & 5.47 & 5.73
        \\
        OmniGen2~\citep{wu2025omnigen2} & 8.05 & 7.58 & 7.11 & 7.13 & 7.45 & 6.38 & 6.71 & 7.04 & 7.18
        \\
        Qwen-Image-Edit-2509~\citep{wu2025qwenimagetechnicalreport} & \textbf{8.35} & \textbf{9.13} & 7.65 & \textbf{8.85} & 7.90 & 5.16 & 7.75 & 6.73 & 7.69
        \\
        DreamOmni2~\cite{xia2025dreamomni2} &7.36 &7.43 &6.10 &6.73 &6.66 &5.20 &5.34 &5.64 &6.31
        \\
        \textbf{\ours{} (Ours)} & 8.25 & 8.55 & \textbf{8.25} & 8.07 & \textbf{8.28} & \textbf{8.21} & \textbf{8.25} & \textbf{7.82} & \textbf{8.21}
        \\
        \bottomrule
    \end{tabular}
}
\caption{Quantitative comparison results on OmniContext~\cite{wu2025omnigen2}. "Char. + Obj." indicates Character + Object.}
\vspace{-5pt}
\label{tab:comp}
\end{table*}

\begin{figure*}[!t]
    \centering
    \includegraphics[width=\linewidth]{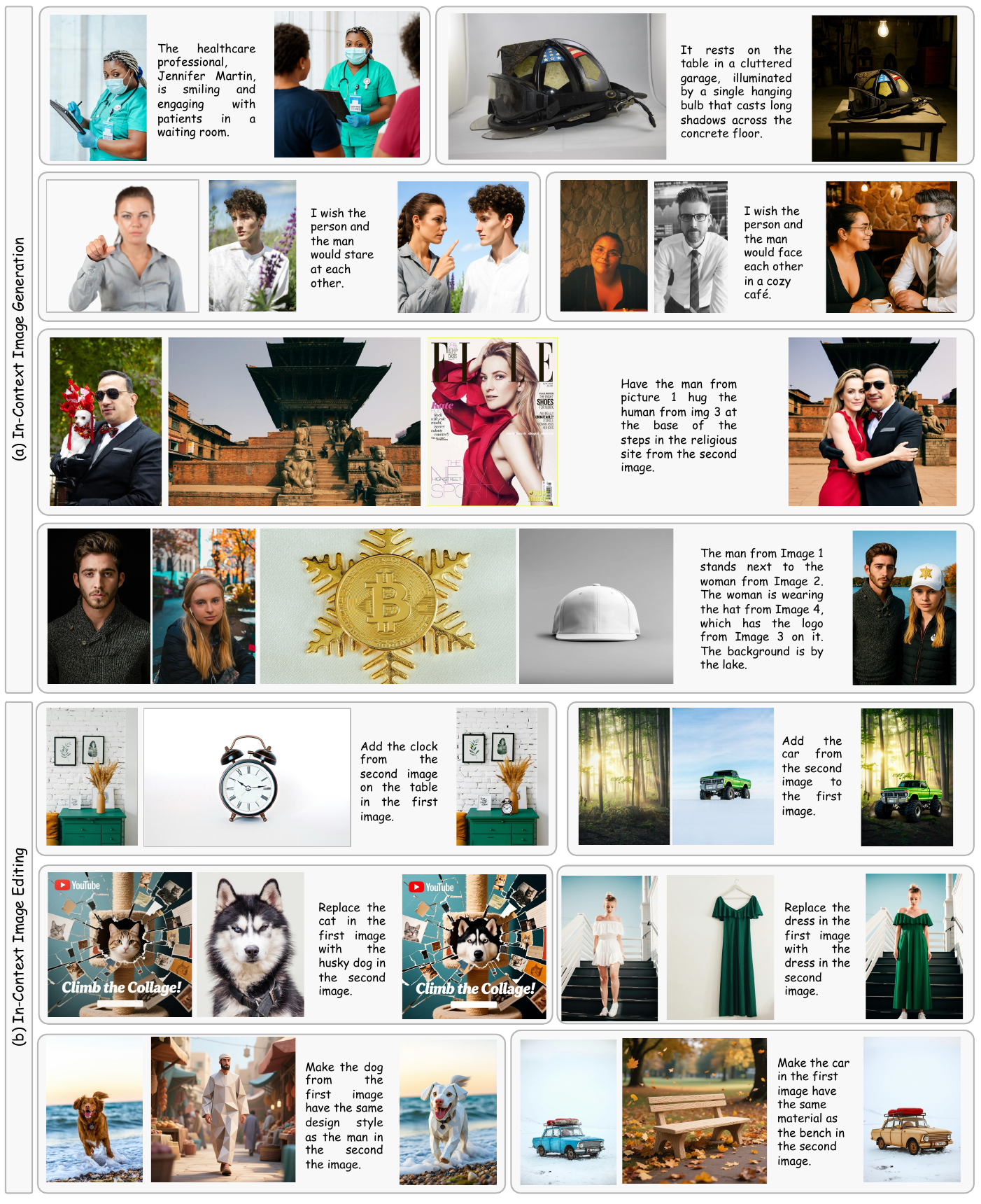}
    \vspace{-6mm}
    \caption{More examples for In-Context Image Generation and Editing. The last image in each group is the generated result, and the others are input reference images. 
    }
    \label{fig:more}
    \vspace{-1mm}
\end{figure*}

\subsection{Qualitative Results}
As shown in Figure~\ref{fig:comp}, we provide qualitative comparisons with recent baselines on the in-context image generation and editing tasks.
For the in-context image generation task in the first two rows, most methods are able to produce roughly correct images. However, OmniGen2~\cite{wu2025omnigen2} often incorporates irrelevant elements from the reference images, such as the instrument in the 1st row and the blue background in the 2nd row. \bagel{}~\cite{deng2025bagel}, Qwen-Image-Edit~\cite{wu2025qwenimagetechnicalreport}, and DreamOmni2~\cite{xia2025dreamomni2} exhibit weaker subject consistency, resulting in a significant mismatch in the appearance of the humans in the 2nd row. For the in-context image editing task in the last three rows, most existing models fail to correctly interpret the editing intent. This is particularly evident in the material replacement shown in the 3rd row and the object addition in the 4th row, where the generated outputs often deviate substantially from the desired edits.
Although DreamOmni2 has almost finished editing the 4th row, there are changes in hand gestures and inconsistencies in background light.
Overall, \ours{} demonstrates a distinct advantage in addressing complex in-context generation and editing challenges.

\begin{table*}[t]
    \centering
    \resizebox{0.99\linewidth}{!}{
    \begin{tabular}{l|cccccccccccc|ccc}
        \toprule
        \multirow{3}{*}{\bf Model} & \multicolumn{12}{c|}{\bf Editing} & \multicolumn{3}{c}{\multirow{2}{*}{\bf Generation}}
        \\ 
        \cmidrule(lr){2-13}
          & \multicolumn{3}{c|}{\bf Add} & \multicolumn{3}{c|}{\bf Replace} & \multicolumn{3}{c|}{\bf Global} & \multicolumn{3}{c|}{\bf Local}  &    &    
        \\
        \cmidrule(lr){2-16}
          & PF & SC & \multicolumn{1}{c|}{Overall} & PF & SC & \multicolumn{1}{c|}{Overall}& PF & SC & \multicolumn{1}{c|}{Overall}& PF & SC & \multicolumn{1}{c|}{Overall} & PF & SC & Overall 
        \\
        \midrule
\bagel{}~\cite{deng2025bagel} & 4.09 & 5.18 & 4.58 &        1.15 & 5.48 & 1.37 &        2.09 & 4.98 & 2.46 &       1.61 & 3.58 & 1.63 &                   5.72 & 5.77 & 5.25 \\
OmniGen2~\cite{wu2025omnigen2} & 7.64 & 7.55 & 7.52 &        5.37 & 6.07 & 5.6 &         6.81 & 7.28 & 6.88 &       2.76 & 5.28 & 2.99 &                  5.15 & 5.74 & 4.99 \\
Echo-4o~\cite{ye2025echo} & 8.36 & 8.73 & 8.51 &        3.85 & 6.52 & 4.51 &        4.38 & 7.66 & 5.16 &       1.92 & 6.24 & 2.41 &                   6.68 & 7.2 & 6.59 \\
Qwen-Image-Edit(2509)~\cite{wu2025qwenimagetechnicalreport} & 6.09 & 7.91 & 6.51 &        2.48 & 4.93 & 2.79 &        3.26 & 5.28 & 3.21 &       2.49 & 4.98 & 2.73 &                   5.98 & 5.78 & 5.45 \\
DreamOmni2*~\cite{xia2025dreamomni2} & 6.73 & 7.91 & 6.87 &        6.78 & 7.56 & 7.05 &        7.34 & 8.68 & 7.76 &       5.14 & 8.18 & 5.44 &                   7.01 & 6.71 & 6.56 \\
\textbf{\ours{} (Ours)} & \textbf{9.27} & \textbf{9.27} & \textbf{9.27} &        \textbf{8.44} & \textbf{8.81} & \textbf{8.61} &        \textbf{7.47} & \textbf{8.57} & \textbf{7.85} &       \textbf{6.11} & \textbf{8.54} & \textbf{6.35} &                   \textbf{7.74} & \textbf{7.67} & \textbf{7.24} \\
        \bottomrule
    \end{tabular}
}
\caption{Quantitative comparison results on DreamOmni2Bench~\cite{xia2025dreamomni2}. Prompt Following (PF), Subject
Consistency (SC), and Overall scores are reported (higher is better). * denotes that DreamOmni2 employs different parameters for editing and generation tasks.}
\label{tab:comp_dreamomni2}
\end{table*}

\begin{table*}[t]
    \centering
    \resizebox{0.99\linewidth}{!}{
    \begin{tabular}{l|ccc|cccccccccccc}
        \toprule
        \multirow{3}{*}{\bf Model} & \multicolumn{3}{c|}{{\bf Editing}} & \multicolumn{12}{c}{\bf Generation} 
        \\ 
        \cmidrule(lr){2-16}
          & \multicolumn{3}{c|}{\bf 2} & \multicolumn{3}{c|}{\bf 1} & \multicolumn{3}{c|}{\bf 2} & \multicolumn{3}{c|}{\bf 3} & \multicolumn{3}{c}{\bf 4}  
        \\
        \cmidrule(lr){2-16}
          & PF & SC & \multicolumn{1}{c|}{Overall} & PF & SC & \multicolumn{1}{c|}{Overall}& PF & SC & \multicolumn{1}{c|}{Overall}& PF & SC & \multicolumn{1}{c|}{Overall} & PF & SC & Overall 
        \\
        \midrule

BAGEL~\cite{deng2025bagel} & 1.8 & 4.28 & 1.97 & 6.38 & 5.49 & 4.76 & 4.92 & 5.52 & 5.04 & 5.28 & 6.03 & 5.52 & 6.05 & 6.24 & 6.04 \\
Echo-4o~\cite{ye2025echo} & 3.16 & 6.78 & 3.72 & \underline{7.44} & \underline{6.79} & \textbf{6.68} & 4.8 & 6.84 & 5.1 & 6.59 & 7.52 & 6.9 & \textbf{7.67} & \textbf{7.95} & \textbf{7.75} \\
OmniGen2~\cite{wu2025omnigen2} & 4.41 & 6.02 & 4.58 & 5.77 & 4.97 & 4.45 & 4.64 & 6.64 & 5.34 & 5.17 & 6.0 & 5.42 & 4.57 & 5.71 & 5.0 \\
Qwen-Image-Edit(2509)~\cite{wu2025qwenimagetechnicalreport} & 2.88 & 5.21 & 3.06 & 7.33 & 5.36 & 5.34 & 5.28 & 6.4 & 5.65 & 5.72 & 6.1 & 5.76 & 4.67 & 5.38 & 4.95 \\
DreamOmni2*~\cite{xia2025dreamomni2} & \underline{6.01} & \underline{8.21} & \underline{6.33} & 6.97 & 5.74 & 5.69 & \underline{7.52} & \textbf{8.2} & \underline{7.65} & \underline{7.0} & \underline{6.62} & \underline{6.76} & 6.48 & 6.86 & 6.61 \\
\textbf{Re-Align (Ours)} & \textbf{6.94} & \textbf{8.62} & \textbf{7.19} & \textbf{7.92} & \textbf{7.1} & \underline{6.37} & \textbf{8.04} & \underline{8.04} & \textbf{7.93} & \textbf{7.72} & \textbf{7.93} & \textbf{7.7 }& \underline{7.05} & \underline{7.9} & \underline{7.39} \\
        \bottomrule
    \end{tabular}
}
\caption{Impact of reference image number on DreamOmni2Bench~\cite{xia2025dreamomni2}.}
\label{tab:comp_dreamomni2_num}
\end{table*}

\subsection{Quantitative Results}
We present quantitative comparisons for in-context image generation on the OmniContext benchmark~\cite{wu2025omnigen2}, as reported in Table~\ref{tab:comp}, and for both in-context image editing and generation on DreamOmni2Bench~\cite{xia2025dreamomni2}, as reported in Table~\ref{tab:comp_dreamomni2}.
Compared with models having comparable scale and computational resources, \ours{} achieves the highest overall average score (Table~\ref{tab:comp}). 
It ranks second only to Qwen-Image-Edit~\cite{wu2025qwenimagetechnicalreport} in the SINGLE task and achieves the best overall performance in MULTIPLE and SCENE tasks, demonstrating the effectiveness of our approach for in-context image generation.
This finding is consistent with the assessment in the generation section of Table~\ref{tab:comp_dreamomni2}. 
The editing section of DreamOmni2Bench~\cite{xia2025dreamomni2} covers Add, Replace, Global, and Local edits, where Add and Replace focus on subject-referenced editing, and Global and Local on attribute-referenced editing. 
Echo-4o~\cite{ye2025echo} performs well in the Add task but poorly in the more complex Global and Local edits. 
DreamOmni2~\cite{xia2025dreamomni2}, which employs separate parameters for generation and editing, exhibits balanced performance across editing types but remains inferior overall to \ours{}. 
In contrast, \ours{} consistently attains higher PF and SC scores across tasks, highlighting its strong advantage in in-context image editing.

\begin{table}[!htbp]
\begin{center}
  \resizebox{\linewidth}{!}{
    \begin{tabular}{c c c| c c c c}
        \toprule
        \multirow{1}{*}{SFT} & \multirow{1}{*}{RGA} & \multirow{1}{*}{RID} & PF~$\uparrow$ & SC~$\uparrow$ & \multicolumn{1}{c}{Overall~$\uparrow$} & \multirow{1}{*}{CLIP$_{out}\uparrow$}\\ 
        \midrule
        \ding{55} &\ding{55}& \ding{55}&  6.92	&5.47	&5.80 &32.44\\
        \ding{51} &\ding{55}& \ding{55}&  \underline{7.51}&	6.46&	6.77 &33.32\\
        \ding{51} & \ding{51} & \ding{55}&  7.46	&\underline{6.54}	&\underline{6.80}	&\underline{33.50}\\
        \ding{51} & \ding{51} & \ding{51} & \textbf{7.61}  &	\textbf{6.57}	&\textbf{6.89} &\textbf{33.90}\\
        \bottomrule
    \end{tabular}
  }
  \caption{Ablation studies on the training stages and strategies. “SFT” denotes supervised fine-tuning for image generation conditioned on IC-CoT, “RGA” represents reasoning–generation alignment, and “RID” refers to the reasoning-induced diversity strategy.}
  \label{tab:abla_cot}
\end{center}
\end{table}

\subsection{Ablation Study}

\begin{figure}[!t]
    \centering
    \includegraphics[width=1\linewidth]{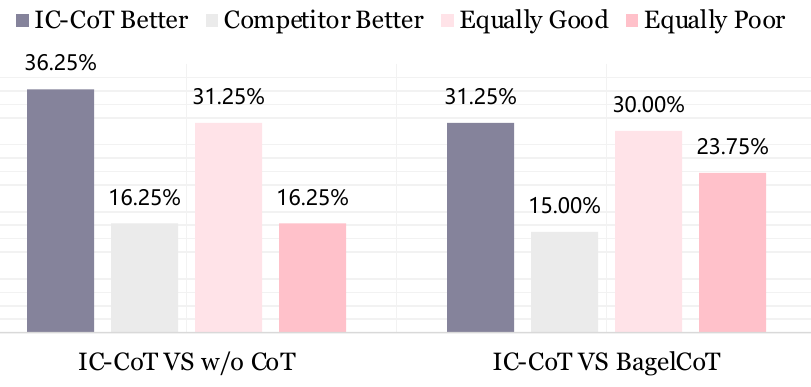}
    \caption{GSB evaluation results from the ablation study on the reasoning mechanism.}
    \vspace{-2mm}
    \label{fig:abla_cot}
\end{figure}

We perform an ablation study to validate the effectiveness of the proposed IC-CoT. 
Specifically, we compare it with two variants: one that excludes the reasoning process (w/o CoT) and another that adopts unstructured reasoning following in~\cite{deng2025bagel} (BagelCoT). 
As illustrated in Figure~\ref{fig:abla_cot}, results from the GSB (Good/Same/Bad) evaluation clearly demonstrate that IC-CoT outperforms the two variants, with win rates 20\% and 16.25\% higher, respectively, confirming the superiority of IC-CoT design.

\begin{figure}[!t]
    \centering
    \includegraphics[width=1\linewidth]{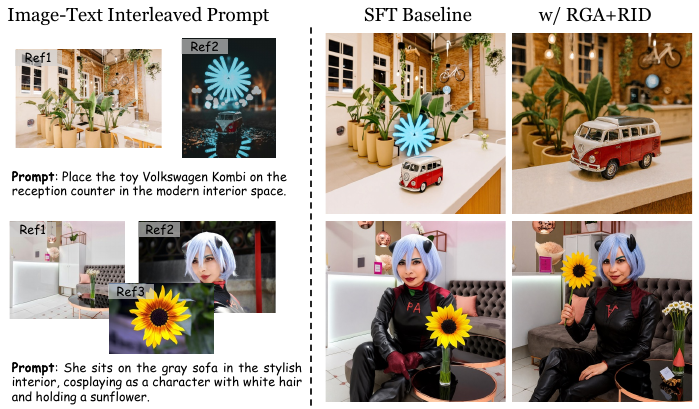}
    \vspace{-5mm}
    \caption{
    Ablation visualization of reasoning–generation alignment with reasoning-induced diversity (RGA+RID).
    }
    \vspace{-3mm}
    \label{fig:abla_vis}
\end{figure}

Besides, we conduct ablation studies to evaluate the effectiveness of the training stages and strategies in \ours{}. 
As shown in Table~\ref{tab:abla_cot}, we report the OmniContextScore~\cite{wu2025omnigen2} along with an additional metric $\text{CLIP}_{out}$ assessing text–image consistency between the generated image and the ground-truth caption on a subset of OmniContext.
After supervised fine-tuning (SFT), the model learns image generation guided by the IC-CoT reasoning, achieving significant improvements across all metrics.
Reasoning–generation alignment (RGA) improves the $\text{CLIP}_{out}$ score but brings no significant gain in PF score, indicating that low sample diversity adversely affects RL training. 
When the reasoning-induced diversity (RID) strategy is subsequently applied, overall performance improves, highlighting the critical role of output diversity in alignment training.
This is consistent with the results shown in Figure~\ref{fig:abla_vis}, where the well-aligned model produces images that better reflect the intended instructions.

\subsection{More Results}
\noindent \textbf{More Visualization}
Figure~\ref{fig:more}(a) provides additional in-context image generation and editing examples, demonstrating that the model produces accurate and highly consistent images when conditioned on one to four reference inputs. 
Furthermore, Figure~\ref{fig:more}(b) showcases in-context image editing capabilities, where the first, second, and third rows illustrate object addition, object replacement, and attribute modification with reference images, respectively. These results underscore the strong versatility and effectiveness of \ours{} across a broad range of creative generation tasks.

\noindent \textbf{Impact of Reference Image Number}
As presented in Table~\ref{tab:comp_dreamomni2_num}, to evaluate the model’s performance with varying numbers of reference images, we conduct experiments on DreamOmni2Bench~\cite{xia2025dreamomni2}.
Two reference images are used for all editing tasks, while generation tasks employ one to four reference images.\ours{} consistently delivers strong PF and SC scores across all configurations, frequently ranking first or second across all metrics and thus demonstrating the most robust overall performance. In contrast, other models struggle to maintain comparable results.

\begin{figure}[!t]
    \centering
    \includegraphics[width=\linewidth]{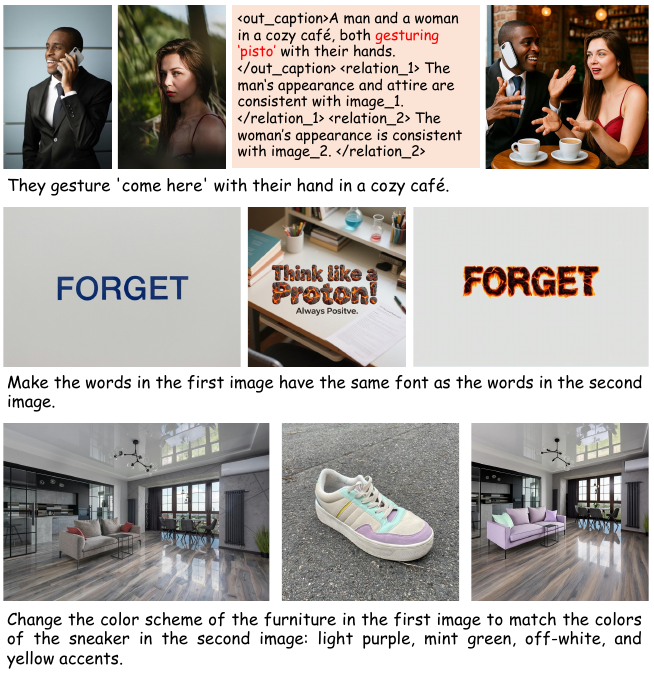}
    \vspace{-6mm}
    \caption{Failure cases of \ours{}. 
    In the first row, the model-generated reasoning text appears on an orange background, with incorrect parts marked in red.
    }
    \vspace{-10pt}
    \label{fig:fail}
    \vspace{-1mm}
\end{figure}

\noindent \textbf{Failure Cases}
We also show several failure cases on the ICGE task as shown in Figure~\ref{fig:fail}.
First, in rare cases, the model fails to generate correct reasoning texts. For example, when processing the complex action semantics of ``\emph{come here}'', it results in subpar image outputs. Furthermore, in editing tasks without dedicated training (e.g., editing based on referenced text styles or object color schemes), the model demonstrates semantic comprehension but produces images with low reference consistency.
Scaling up the model size and integrating more comprehensive training data may help alleviate these issues.

\section{Conclusion and Limitation}
In this work, we propose a unified framework for in-context image generation and editing that bridges understanding and generation via a reasoning mechanism. 
We design an IC-CoT that provides explicit semantic guidance and reference association, providing clear targets for subsequent image generation. 
The reasoning alignment stage further enhances consistency between the reasoning content and the generated image via policy optimization.
Despite these advances, our work still faces several challenges. First, our model size and data scale are limited compared to production-level work like GPT-4o~\cite{gpt4o}, which may constrain the model's performance in diverse scenarios.
Second, the current IC-CoT operates purely at the textual level; extending it to visual Chain-of-Thought reasoning may be a promising direction for future work.


{
    \small
    \bibliographystyle{ieeenat_fullname}
    \bibliography{main}

\begin{thebibliography}{74}
\providecommand{\natexlab}[1]{#1}
\providecommand{\url}[1]{\texttt{#1}}
\expandafter\ifx\csname urlstyle\endcsname\relax
  \providecommand{\doi}[1]{doi: #1}\else
  \providecommand{\doi}{doi: \begingroup \urlstyle{rm}\Url}\fi

\bibitem[Achiam et~al.(2023)Achiam, Adler, Agarwal, Ahmad, Akkaya, Aleman, Almeida, Altenschmidt, Altman, Anadkat, et~al.]{achiam2023gpt4}
Josh Achiam, Steven Adler, Sandhini Agarwal, Lama Ahmad, Ilge Akkaya, Florencia~Leoni Aleman, Diogo Almeida, Janko Altenschmidt, Sam Altman, Shyamal Anadkat, et~al.
\newblock Gpt-4 technical report.
\newblock \emph{arXiv preprint arXiv:2303.08774}, 2023.

\bibitem[Bai et~al.(2025)Bai, Chen, Liu, Wang, Ge, Song, Dang, Wang, Wang, Tang, et~al.]{bai2025qwen25vl}
Shuai Bai, Keqin Chen, Xuejing Liu, Jialin Wang, Wenbin Ge, Sibo Song, Kai Dang, Peng Wang, Shijie Wang, Jun Tang, et~al.
\newblock Qwen2. 5-vl technical report.
\newblock \emph{arXiv preprint arXiv:2502.13923}, 2025.

\bibitem[Black et~al.(2023)Black, Janner, Du, Kostrikov, and Levine]{ddpo}
Kevin Black, Michael Janner, Yilun Du, Ilya Kostrikov, and Sergey Levine.
\newblock Training diffusion models with reinforcement learning.
\newblock \emph{arXiv preprint arXiv:2305.13301}, 2023.

\bibitem[Brooks et~al.(2023)Brooks, Holynski, and Efros]{brooks2023instructpix2pix}
Tim Brooks, Aleksander Holynski, and Alexei~A Efros.
\newblock Instructpix2pix: Learning to follow image editing instructions.
\newblock In \emph{Proceedings of the IEEE/CVF Conference on Computer Vision and Pattern Recognition}, pages 18392--18402, 2023.

\bibitem[Brown et~al.(2020)Brown, Mann, Ryder, Subbiah, Kaplan, Dhariwal, Neelakantan, Shyam, Sastry, Askell, Agarwal, Herbert{-}Voss, Krueger, Henighan, Child, Ramesh, Ziegler, Wu, Winter, Hesse, Chen, Sigler, Litwin, Gray, Chess, Clark, Berner, McCandlish, Radford, Sutskever, and Amodei]{Brown2020GPT3}
Tom~B. Brown, Benjamin Mann, Nick Ryder, Melanie Subbiah, Jared Kaplan, Prafulla Dhariwal, Arvind Neelakantan, Pranav Shyam, Girish Sastry, Amanda Askell, Sandhini Agarwal, Ariel Herbert{-}Voss, Gretchen Krueger, Tom Henighan, Rewon Child, Aditya Ramesh, Daniel~M. Ziegler, Jeffrey Wu, Clemens Winter, Christopher Hesse, Mark Chen, Eric Sigler, Mateusz Litwin, Scott Gray, Benjamin Chess, Jack Clark, Christopher Berner, Sam McCandlish, Alec Radford, Ilya Sutskever, and Dario Amodei.
\newblock Language models are few-shot learners.
\newblock In \emph{Advances in Neural Information Processing Systems 33: Annual Conference on Neural Information Processing Systems 2020, NeurIPS 2020, December 6-12, 2020, virtual}, 2020.

\bibitem[Cao et~al.(2025)Cao, Chen, Chen, Cheng, Cui, Deng, Dong, Gong, Gu, Gu, et~al.]{cao2025hunyuanimage3}
Siyu Cao, Hangting Chen, Peng Chen, Yiji Cheng, Yutao Cui, Xinchi Deng, Ying Dong, Kipper Gong, Tianpeng Gu, Xiusen Gu, et~al.
\newblock Hunyuanimage 3.0 technical report.
\newblock \emph{arXiv preprint arXiv:2509.23951}, 2025.

\bibitem[Chang et~al.(2022)Chang, Zhang, Jiang, Liu, and Freeman]{chang2022maskgit}
Huiwen Chang, Han Zhang, Lu Jiang, Ce Liu, and William~T Freeman.
\newblock Maskgit: Masked generative image transformer.
\newblock In \emph{Proceedings of the IEEE/CVF conference on computer vision and pattern recognition}, pages 11315--11325, 2022.

\bibitem[Chen et~al.(2025{\natexlab{a}})Chen, Cai, Chen, Chen, Ji, Wang, Yang, and Wang]{chen2025sharegpt}
Junying Chen, Zhenyang Cai, Pengcheng Chen, Shunian Chen, Ke Ji, Xidong Wang, Yunjin Yang, and Benyou Wang.
\newblock Sharegpt-4o-image: Aligning multimodal models with gpt-4o-level image generation.
\newblock \emph{arXiv preprint arXiv:2506.18095}, 2025{\natexlab{a}}.

\bibitem[Chen et~al.(2025{\natexlab{b}})Chen, Xu, Pan, Hu, Qin, Goldstein, Huang, Zhou, Xie, Savarese, et~al.]{chen2025blip3o}
Jiuhai Chen, Zhiyang Xu, Xichen Pan, Yushi Hu, Can Qin, Tom Goldstein, Lifu Huang, Tianyi Zhou, Saining Xie, Silvio Savarese, et~al.
\newblock Blip3-o: A family of fully open unified multimodal models-architecture, training and dataset.
\newblock \emph{arXiv preprint arXiv:2505.09568}, 2025{\natexlab{b}}.

\bibitem[Chen et~al.(2025{\natexlab{c}})Chen, Wu, Liu, Pan, Liu, Xie, Yu, and Ruan]{chen2025janus_pro}
Xiaokang Chen, Zhiyu Wu, Xingchao Liu, Zizheng Pan, Wen Liu, Zhenda Xie, Xingkai Yu, and Chong Ruan.
\newblock Janus-pro: Unified multimodal understanding and generation with data and model scaling.
\newblock \emph{arXiv preprint arXiv:2501.17811}, 2025{\natexlab{c}}.

\bibitem[Cui et~al.(2025)Cui, Chen, Deng, Huang, Li, Liu, Liu, Luo, Wang, Wang, et~al.]{cui2025emu35}
Yufeng Cui, Honghao Chen, Haoge Deng, Xu Huang, Xinghang Li, Jirong Liu, Yang Liu, Zhuoyan Luo, Jinsheng Wang, Wenxuan Wang, et~al.
\newblock Emu3. 5: Native multimodal models are world learners.
\newblock \emph{arXiv preprint arXiv:2510.26583}, 2025.

\bibitem[Deng et~al.(2025)Deng, Zhu, Li, Gou, Li, Wang, Zhong, Yu, Nie, Song, et~al.]{deng2025bagel}
Chaorui Deng, Deyao Zhu, Kunchang Li, Chenhui Gou, Feng Li, Zeyu Wang, Shu Zhong, Weihao Yu, Xiaonan Nie, Ziang Song, et~al.
\newblock Emerging properties in unified multimodal pretraining.
\newblock \emph{arXiv preprint arXiv:2505.14683}, 2025.

\bibitem[Dubey et~al.(2024)Dubey, Jauhri, Pandey, Kadian, Al-Dahle, Letman, Mathur, Schelten, Yang, Fan, et~al.]{dubey2024llama3.1-paper}
Abhimanyu Dubey, Abhinav Jauhri, Abhinav Pandey, Abhishek Kadian, Ahmad Al-Dahle, Aiesha Letman, Akhil Mathur, Alan Schelten, Amy Yang, Angela Fan, et~al.
\newblock The llama 3 herd of models.
\newblock \emph{arXiv preprint arXiv:2407.21783}, 2024.

\bibitem[Fan et~al.(2023)Fan, Watkins, Du, Liu, Ryu, Boutilier, Abbeel, Ghavamzadeh, Lee, and Lee]{fan2023dpok}
Ying Fan, Olivia Watkins, Yuqing Du, Hao Liu, Moonkyung Ryu, Craig Boutilier, Pieter Abbeel, Mohammad Ghavamzadeh, Kangwook Lee, and Kimin Lee.
\newblock Dpok: Reinforcement learning for fine-tuning text-to-image diffusion models.
\newblock \emph{Advances in Neural Information Processing Systems}, 36:\penalty0 79858--79885, 2023.

\bibitem[Fang et~al.(2025)Fang, Duan, Wang, Huang, Li, Yan, Tian, Zeng, Zhao, Dai, et~al.]{fang2025got}
Rongyao Fang, Chengqi Duan, Kun Wang, Linjiang Huang, Hao Li, Shilin Yan, Hao Tian, Xingyu Zeng, Rui Zhao, Jifeng Dai, et~al.
\newblock Got: Unleashing reasoning capability of multimodal large language model for visual generation and editing.
\newblock \emph{arXiv preprint arXiv:2503.10639}, 2025.

\bibitem[Gal et~al.(2022{\natexlab{a}})Gal, Alaluf, Atzmon, Patashnik, Bermano, Chechik, and Cohen-Or]{gal2022ti}
Rinon Gal, Yuval Alaluf, Yuval Atzmon, Or Patashnik, Amit~H. Bermano, Gal Chechik, and Daniel Cohen-Or.
\newblock An image is worth one word: Personalizing text-to-image generation using textual inversion, 2022{\natexlab{a}}.

\bibitem[Gal et~al.(2022{\natexlab{b}})Gal, Patashnik, Maron, Bermano, Chechik, and Cohen-Or]{gal2022stylegan}
Rinon Gal, Or Patashnik, Haggai Maron, Amit~H Bermano, Gal Chechik, and Daniel Cohen-Or.
\newblock Stylegan-nada: Clip-guided domain adaptation of image generators.
\newblock \emph{ACM Transactions on Graphics (TOG)}, 41\penalty0 (4):\penalty0 1--13, 2022{\natexlab{b}}.

\bibitem[Google(2025)]{gemini-2.0-flash}
Google.
\newblock Gemini 2.0 flash.
\newblock \url{https://developers.googleblog.com/en/experiment-with-gemini-20-flash-native-image-generation}, 2025.

\bibitem[{Google}(2025)]{google2025nanobanana}
{Google}.
\newblock Nano banana.
\newblock Technical report, Google, 2025.

\bibitem[Guo et~al.(2025)Guo, Yang, Zhang, Song, Zhang, Xu, Zhu, Ma, Wang, Bi, et~al.]{guo2025deepseekr1}
Daya Guo, Dejian Yang, Haowei Zhang, Junxiao Song, Ruoyu Zhang, Runxin Xu, Qihao Zhu, Shirong Ma, Peiyi Wang, Xiao Bi, et~al.
\newblock Deepseek-r1: Incentivizing reasoning capability in llms via reinforcement learning.
\newblock \emph{arXiv preprint arXiv:2501.12948}, 2025.

\bibitem[He et~al.(2024)He, Ma, Huang, Huang, Gao, Wei, Dai, Han, and Liu]{he2024freeedit}
Runze He, Kai Ma, Linjiang Huang, Shaofei Huang, Jialin Gao, Xiaoming Wei, Jiao Dai, Jizhong Han, and Si Liu.
\newblock Freeedit: Mask-free reference-based image editing with multi-modal instruction, 2024.

\bibitem[He et~al.(2025)He, Cheng, Ma, Jia, Liu, Ma, Wu, Wu, Leng, and Yin]{he2025plangen}
Runze He, Bo Cheng, Yuhang Ma, Qingxiang Jia, Shanyuan Liu, Ao Ma, Xiaoyu Wu, Liebucha Wu, Dawei Leng, and Yuhui Yin.
\newblock Plangen: Towards unified layout planning and image generation in auto-regressive vision language models.
\newblock \emph{arXiv preprint arXiv:2503.10127}, 2025.

\bibitem[Ho et~al.(2020)Ho, Jain, and Abbeel]{ho2020ddpm}
Jonathan Ho, Ajay Jain, and Pieter Abbeel.
\newblock Denoising diffusion probabilistic models.
\newblock In \emph{NeurIPS}, 2020.

\bibitem[Hui et~al.(2024)Hui, Yang, Cui, Yang, Liu, Zhang, Liu, Zhang, Yu, Dang, et~al.]{hui2024qwen2.5-paper}
Binyuan Hui, Jian Yang, Zeyu Cui, Jiaxi Yang, Dayiheng Liu, Lei Zhang, Tianyu Liu, Jiajun Zhang, Bowen Yu, Kai Dang, et~al.
\newblock Qwen2. 5-coder technical report.
\newblock \emph{arXiv preprint arXiv:2409.12186}, 2024.

\bibitem[Isola et~al.(2017)Isola, Zhu, Zhou, and Efros]{Isola2017pix2pix}
Phillip Isola, Jun{-}Yan Zhu, Tinghui Zhou, and Alexei~A. Efros.
\newblock Image-to-image translation with conditional adversarial networks.
\newblock In \emph{2017 {IEEE} Conference on Computer Vision and Pattern Recognition, {CVPR} 2017, Honolulu, HI, USA, July 21-26, 2017}, pages 5967--5976. {IEEE} Computer Society, 2017.

\bibitem[Jiang et~al.(2025)Jiang, Yan, Jia, Liu, Kang, and Lu]{jiang2025infiniteyou}
Liming Jiang, Qing Yan, Yumin Jia, Zichuan Liu, Hao Kang, and Xin Lu.
\newblock Infiniteyou: Flexible photo recrafting while preserving your identity.
\newblock \emph{arXiv preprint arXiv:2503.16418}, 2025.

\bibitem[Kirstain et~al.(2023)Kirstain, Polyak, Singer, Matiana, Penna, and Levy]{kirstain2023pick}
Yuval Kirstain, Adam Polyak, Uriel Singer, Shahbuland Matiana, Joe Penna, and Omer Levy.
\newblock Pick-a-pic: An open dataset of user preferences for text-to-image generation.
\newblock \emph{arXiv preprint arXiv:2305.01569}, 2023.

\bibitem[Ku et~al.(2023)Ku, Jiang, Wei, Yue, and Chen]{ku2023viescore}
Max Ku, Dongfu Jiang, Cong Wei, Xiang Yue, and Wenhu Chen.
\newblock Viescore: Towards explainable metrics for conditional image synthesis evaluation.
\newblock \emph{arXiv preprint arXiv:2312.14867}, 2023.

\bibitem[Kumari et~al.(2023)Kumari, Zhang, Zhang, Shechtman, and Zhu]{kumari2022customdiffusion}
Nupur Kumari, Bingliang Zhang, Richard Zhang, Eli Shechtman, and Jun-Yan Zhu.
\newblock Multi-concept customization of text-to-image diffusion.
\newblock 2023.

\bibitem[Labs(2024)]{flux}
Black~Forest Labs.
\newblock Flux.
\newblock \url{https://blackforestlabs.ai/announcing-black-forest-labs}, 2024.

\bibitem[Labs et~al.(2025)Labs, Batifol, Blattmann, Boesel, Consul, Diagne, Dockhorn, English, English, Esser, et~al.]{labs2025flux}
Black~Forest Labs, Stephen Batifol, Andreas Blattmann, Frederic Boesel, Saksham Consul, Cyril Diagne, Tim Dockhorn, Jack English, Zion English, Patrick Esser, et~al.
\newblock Flux. 1 kontext: Flow matching for in-context image generation and editing in latent space.
\newblock \emph{arXiv preprint arXiv:2506.15742}, 2025.

\bibitem[LAION(2022)]{LAION-Aesthetics}
LAION.
\newblock Laion-aesthetics v2.
\newblock \url{https://github.com/christophschuhmann/improved-aesthetic-predictor}, 2022.

\bibitem[Li et~al.(2023)Li, Li, and Hoi]{li2023blipdf}
Dongxu Li, Junnan Li, and Steven~CH Hoi.
\newblock Blip-diffusion: Pre-trained subject representation for controllable text-to-image generation and editing.
\newblock \emph{arXiv preprint arXiv:2305.14720}, 2023.

\bibitem[Liu et~al.(2025)Liu, Liu, Liang, Li, Liu, Wang, Wan, Zhang, and Ouyang]{liu2025flowgrpo}
Jie Liu, Gongye Liu, Jiajun Liang, Yangguang Li, Jiaheng Liu, Xintao Wang, Pengfei Wan, Di Zhang, and Wanli Ouyang.
\newblock Flow-grpo: Training flow matching models via online rl.
\newblock \emph{arXiv preprint arXiv:2505.05470}, 2025.

\bibitem[Liu et~al.(2022)Liu, Gong, and Liu]{liu2022flowmatching}
Xingchao Liu, Chengyue Gong, and Qiang Liu.
\newblock Flow straight and fast: Learning to generate and transfer data with rectified flow.
\newblock \emph{arXiv preprint arXiv:2209.03003}, 2022.

\bibitem[Ma et~al.(2024)Ma, Liu, Chen, Liu, Wu, Wu, Pan, Xie, Zhang, yu, Zhao, Wang, Liu, and Ruan]{ma2024janusflow}
Yiyang Ma, Xingchao Liu, Xiaokang Chen, Wen Liu, Chengyue Wu, Zhiyu Wu, Zizheng Pan, Zhenda Xie, Haowei Zhang, Xingkai yu, Liang Zhao, Yisong Wang, Jiaying Liu, and Chong Ruan.
\newblock Janusflow: Harmonizing autoregression and rectified flow for unified multimodal understanding and generation, 2024.

\bibitem[Mo et~al.(2024)Mo, Mu, Lin, Liu, Guan, Li, and Zhou]{mo2024freecontrol}
Sicheng Mo, Fangzhou Mu, Kuan~Heng Lin, Yanli Liu, Bochen Guan, Yin Li, and Bolei Zhou.
\newblock Freecontrol: Training-free spatial control of any text-to-image diffusion model with any condition.
\newblock In \emph{Proceedings of the IEEE/CVF Conference on Computer Vision and Pattern Recognition}, pages 7465--7475, 2024.

\bibitem[OpenAI(2025{\natexlab{a}})]{gpt4-1}
OpenAI.
\newblock Gpt-4-1.
\newblock \url{https://openai.com/index/gpt-4-1}, 2025{\natexlab{a}}.

\bibitem[OpenAI(2025{\natexlab{b}})]{gpt4o}
OpenAI.
\newblock Gpt-4o.
\newblock \url{https://openai.com/index/introducing-4o-image-generation}, 2025{\natexlab{b}}.

\bibitem[{OpenAI}(2025)]{openai2025gptimage}
{OpenAI}.
\newblock Gpt-image.
\newblock Technical report, OpenAI, 2025.

\bibitem[Pan et~al.(2025)Pan, Shukla, Singh, Zhao, Mishra, Wang, Xu, Chen, Li, Juefei-Xu, et~al.]{pan2025metaquery}
Xichen Pan, Satya~Narayan Shukla, Aashu Singh, Zhuokai Zhao, Shlok~Kumar Mishra, Jialiang Wang, Zhiyang Xu, Jiuhai Chen, Kunpeng Li, Felix Juefei-Xu, et~al.
\newblock Transfer between modalities with metaqueries.
\newblock \emph{arXiv preprint arXiv:2504.06256}, 2025.

\bibitem[Peebles and Xie(2023)]{peebles2023dit}
William Peebles and Saining Xie.
\newblock Scalable diffusion models with transformers.
\newblock In \emph{Proceedings of the IEEE/CVF international conference on computer vision}, pages 4195--4205, 2023.

\bibitem[Radford et~al.(2021)Radford, Kim, Hallacy, Ramesh, Goh, Agarwal, Sastry, Askell, Mishkin, Clark, et~al.]{radford2021clip}
Alec Radford, Jong~Wook Kim, Chris Hallacy, Aditya Ramesh, Gabriel Goh, Sandhini Agarwal, Girish Sastry, Amanda Askell, Pamela Mishkin, Jack Clark, et~al.
\newblock Learning transferable visual models from natural language supervision.
\newblock In \emph{ICML}, 2021.

\bibitem[Rombach et~al.(2022)Rombach, Blattmann, Lorenz, Esser, and Ommer]{rombach2022stablediffusion}
Robin Rombach, Andreas Blattmann, Dominik Lorenz, Patrick Esser, and Bj{\"o}rn Ommer.
\newblock High-resolution image synthesis with latent diffusion models.
\newblock In \emph{CVPR}, 2022.

\bibitem[Ruiz et~al.(2022)Ruiz, Li, Jampani, Pritch, Rubinstein, and Aberman]{ruiz2022dreambooth}
Nataniel Ruiz, Yuanzhen Li, Varun Jampani, Yael Pritch, Michael Rubinstein, and Kfir Aberman.
\newblock Dreambooth: Fine tuning text-to-image diffusion models for subject-driven generation.
\newblock \emph{arXiv preprint arXiv:2208.12242}, 2022.

\bibitem[Schulman et~al.(2017)Schulman, Wolski, Dhariwal, Radford, and Klimov]{schulman2017ppo}
John Schulman, Filip Wolski, Prafulla Dhariwal, Alec Radford, and Oleg Klimov.
\newblock Proximal policy optimization algorithms.
\newblock \emph{arXiv preprint arXiv:1707.06347}, 2017.

\bibitem[Shao et~al.(2024)Shao, Wang, Zhu, Xu, Song, Bi, Zhang, Zhang, Li, Wu, et~al.]{shao2024deepseekmath}
Zhihong Shao, Peiyi Wang, Qihao Zhu, Runxin Xu, Junxiao Song, Xiao Bi, Haowei Zhang, Mingchuan Zhang, YK Li, Yang Wu, et~al.
\newblock Deepseekmath: Pushing the limits of mathematical reasoning in open language models.
\newblock \emph{arXiv preprint arXiv:2402.03300}, 2024.

\bibitem[Shen et~al.(2025)Shen, Li, Yang, Zhang, Zhang, Li, Wang, Lu, and Tang]{shen2025srpo}
Xiangwei Shen, Zhimin Li, Zhantao Yang, Shiyi Zhang, Yingfang Zhang, Donghao Li, Chunyu Wang, Qinglin Lu, and Yansong Tang.
\newblock Directly aligning the full diffusion trajectory with fine-grained human preference.
\newblock \emph{arXiv preprint arXiv:2509.06942}, 2025.

\bibitem[Sheynin et~al.(2023)Sheynin, Polyak, Singer, Kirstain, Zohar, Ashual, Parikh, and Taigman]{Sheynin2023emu_edit}
Shelly Sheynin, Adam Polyak, Uriel Singer, Yuval Kirstain, Amit Zohar, Oron Ashual, Devi Parikh, and Yaniv Taigman.
\newblock Emu edit: Precise image editing via recognition and generation tasks.
\newblock 2023.

\bibitem[Song et~al.(2021)Song, Meng, and Ermon]{song2021ddim}
Jiaming Song, Chenlin Meng, and Stefano Ermon.
\newblock Denoising diffusion implicit models.
\newblock In \emph{ICLR}, 2021.

\bibitem[Sun et~al.(2024)Sun, Jiang, Chen, Zhang, Peng, Luo, and Yuan]{sun2024autoregressive_llamagen}
Peize Sun, Yi Jiang, Shoufa Chen, Shilong Zhang, Bingyue Peng, Ping Luo, and Zehuan Yuan.
\newblock Autoregressive model beats diffusion: Llama for scalable image generation.
\newblock \emph{arXiv preprint arXiv:2406.06525}, 2024.

\bibitem[Tao et~al.(2025)Tao, Zhang, Wang, Cheng, Wang, Bai, Zhou, Li, Wang, Wang, et~al.]{tao2025instantcharacter}
Jiale Tao, Yanbing Zhang, Qixun Wang, Yiji Cheng, Haofan Wang, Xu Bai, Zhengguang Zhou, Ruihuang Li, Linqing Wang, Chunyu Wang, et~al.
\newblock Instantcharacter: Personalize any characters with a scalable diffusion transformer framework.
\newblock \emph{arXiv preprint arXiv:2504.12395}, 2025.

\bibitem[Team(2025)]{gemini2p5}
Gemini Team.
\newblock Gemini 2.5: Pushing the frontier with advanced reasoning, multimodality, long context, and next generation agentic capabilities.
\newblock \emph{arXiv preprint arXiv:2507.06261}, 2025.

\bibitem[Tian et~al.(2024)Tian, Jiang, Yuan, Peng, and Wang]{VAR}
Keyu Tian, Yi Jiang, Zehuan Yuan, Bingyue Peng, and Liwei Wang.
\newblock Visual autoregressive modeling: Scalable image generation via next-scale prediction.
\newblock 2024.

\bibitem[Wallace et~al.(2024)Wallace, Dang, Rafailov, Zhou, Lou, Purushwalkam, Ermon, Xiong, Joty, and Naik]{wallace2024dpo}
Bram Wallace, Meihua Dang, Rafael Rafailov, Linqi Zhou, Aaron Lou, Senthil Purushwalkam, Stefano Ermon, Caiming Xiong, Shafiq Joty, and Nikhil Naik.
\newblock Diffusion model alignment using direct preference optimization.
\newblock In \emph{Proceedings of the IEEE/CVF Conference on Computer Vision and Pattern Recognition}, pages 8228--8238, 2024.

\bibitem[Wang et~al.(2024)Wang, Zhang, Luo, Sun, Cui, Wang, Zhang, Wang, Li, Yu, et~al.]{wang2024emu3}
Xinlong Wang, Xiaosong Zhang, Zhengxiong Luo, Quan Sun, Yufeng Cui, Jinsheng Wang, Fan Zhang, Yueze Wang, Zhen Li, Qiying Yu, et~al.
\newblock Emu3: Next-token prediction is all you need.
\newblock \emph{arXiv preprint arXiv:2409.18869}, 2024.

\bibitem[Wang et~al.(2025)Wang, Liu, He, Zhang, Huang, Zhang, Shu, Tao, She, Yu, et~al.]{wang2025mint}
Yi Wang, Mushui Liu, Wanggui He, Longxiang Zhang, Ziwei Huang, Guanghao Zhang, Fangxun Shu, Zhong Tao, Dong She, Zhelun Yu, et~al.
\newblock Mint: Multi-modal chain of thought in unified generative models for enhanced image generation.
\newblock \emph{arXiv preprint arXiv:2503.01298}, 2025.

\bibitem[Wu et~al.(2024)Wu, Chen, Wu, Ma, Liu, Pan, Liu, Xie, Yu, Ruan, et~al.]{wu2024janus}
Chengyue Wu, Xiaokang Chen, Zhiyu Wu, Yiyang Ma, Xingchao Liu, Zizheng Pan, Wen Liu, Zhenda Xie, Xingkai Yu, Chong Ruan, et~al.
\newblock Janus: Decoupling visual encoding for unified multimodal understanding and generation.
\newblock \emph{arXiv preprint arXiv:2410.13848}, 2024.

\bibitem[Wu et~al.(2025{\natexlab{a}})Wu, Li, Zhou, Lin, Gao, Yan, ming Yin, Bai, Xu, Chen, Chen, Tang, Zhang, Wang, Yang, Yu, Cheng, Liu, Li, Zhang, Meng, Wei, Ni, Chen, Cao, Peng, Qu, Wu, Wang, Yu, Wen, Feng, Xu, Wang, Zhang, Zhu, Wu, Cai, and Liu]{wu2025qwenimagetechnicalreport}
Chenfei Wu, Jiahao Li, Jingren Zhou, Junyang Lin, Kaiyuan Gao, Kun Yan, Sheng ming Yin, Shuai Bai, Xiao Xu, Yilei Chen, Yuxiang Chen, Zecheng Tang, Zekai Zhang, Zhengyi Wang, An Yang, Bowen Yu, Chen Cheng, Dayiheng Liu, Deqing Li, Hang Zhang, Hao Meng, Hu Wei, Jingyuan Ni, Kai Chen, Kuan Cao, Liang Peng, Lin Qu, Minggang Wu, Peng Wang, Shuting Yu, Tingkun Wen, Wensen Feng, Xiaoxiao Xu, Yi Wang, Yichang Zhang, Yongqiang Zhu, Yujia Wu, Yuxuan Cai, and Zenan Liu.
\newblock Qwen-image technical report, 2025{\natexlab{a}}.

\bibitem[Wu et~al.(2025{\natexlab{b}})Wu, Zheng, Yan, Xiao, Luo, Wang, Li, Jiang, Liu, Zhou, et~al.]{wu2025omnigen2}
Chenyuan Wu, Pengfei Zheng, Ruiran Yan, Shitao Xiao, Xin Luo, Yueze Wang, Wanli Li, Xiyan Jiang, Yexin Liu, Junjie Zhou, et~al.
\newblock Omnigen2: Exploration to advanced multimodal generation.
\newblock \emph{arXiv preprint arXiv:2506.18871}, 2025{\natexlab{b}}.

\bibitem[Wu et~al.(2025{\natexlab{c}})Wu, Huang, Wu, Cheng, Ding, and He]{uno}
Shaojin Wu, Mengqi Huang, Wenxu Wu, Yufeng Cheng, Fei Ding, and Qian He.
\newblock Less-to-more generalization: Unlocking more controllability by in-context generation.
\newblock \emph{arXiv preprint arXiv:2504.02160}, 2025{\natexlab{c}}.

\bibitem[Xia et~al.(2025)Xia, Peng, Zhang, Huang, Liu, Li, Tan, Wu, Wang, Wang, et~al.]{xia2025dreamomni2}
Bin Xia, Bohao Peng, Yuechen Zhang, Junjia Huang, Jiyang Liu, Jingyao Li, Haoru Tan, Sitong Wu, Chengyao Wang, Yitong Wang, et~al.
\newblock Dreamomni2: Multimodal instruction-based editing and generation.
\newblock \emph{arXiv preprint arXiv:2510.06679}, 2025.

\bibitem[Xiao et~al.(2025)Xiao, Wang, Zhou, Yuan, Xing, Yan, Li, Wang, Huang, and Liu]{xiao2025omnigen}
Shitao Xiao, Yueze Wang, Junjie Zhou, Huaying Yuan, Xingrun Xing, Ruiran Yan, Chaofan Li, Shuting Wang, Tiejun Huang, and Zheng Liu.
\newblock Omnigen: Unified image generation.
\newblock In \emph{Proceedings of the Computer Vision and Pattern Recognition Conference}, pages 13294--13304, 2025.

\bibitem[Xie et~al.(2024)Xie, Mao, Bai, Zhang, Wang, Lin, Gu, Chen, Yang, and Shou]{xie2024showo}
Jinheng Xie, Weijia Mao, Zechen Bai, David~Junhao Zhang, Weihao Wang, Kevin~Qinghong Lin, Yuchao Gu, Zhijie Chen, Zhenheng Yang, and Mike~Zheng Shou.
\newblock Show-o: One single transformer to unify multimodal understanding and generation.
\newblock \emph{arXiv preprint arXiv:2408.12528}, 2024.

\bibitem[Xie et~al.(2025)Xie, Yang, and Shou]{xie2025showo2}
Jinheng Xie, Zhenheng Yang, and Mike~Zheng Shou.
\newblock Show-o2: Improved native unified multimodal models.
\newblock \emph{arXiv preprint arXiv:2506.15564}, 2025.

\bibitem[Xu et~al.(2023)Xu, Liu, Wu, Tong, Li, Ding, Tang, and Dong]{xu2023imagereward}
Jiazheng Xu, Xiao Liu, Yuchen Wu, Yuxuan Tong, Qinkai Li, Ming Ding, Jie Tang, and Yuxiao Dong.
\newblock Imagereward: Learning and evaluating human preferences for text-to-image generation.
\newblock In \emph{NeurIPS}, 2023.

\bibitem[Xue et~al.(2025)Xue, Wu, Gao, Kong, Zhu, Chen, Liu, Liu, Guo, Huang, et~al.]{xue2025dancegrpo}
Zeyue Xue, Jie Wu, Yu Gao, Fangyuan Kong, Lingting Zhu, Mengzhao Chen, Zhiheng Liu, Wei Liu, Qiushan Guo, Weilin Huang, et~al.
\newblock Dancegrpo: Unleashing grpo on visual generation.
\newblock \emph{arXiv preprint arXiv:2505.07818}, 2025.

\bibitem[Ye et~al.(2023)Ye, Zhang, Liu, Han, and Yang]{ye2023ipadapter}
Hu Ye, Jun Zhang, Sibo Liu, Xiao Han, and Wei Yang.
\newblock Ip-adapter: Text compatible image prompt adapter for text-to-image diffusion models.
\newblock \emph{arXiv preprint arXiv:2308.06721}, 2023.

\bibitem[Ye et~al.(2025)Ye, Jiang, Wang, Zhu, Hu, Huang, He, Yan, Yu, Li, et~al.]{ye2025echo}
Junyan Ye, Dongzhi Jiang, Zihao Wang, Leqi Zhu, Zhenghao Hu, Zilong Huang, Jun He, Zhiyuan Yan, Jinghua Yu, Hongsheng Li, et~al.
\newblock Echo-4o: Harnessing the power of gpt-4o synthetic images for improved image generation.
\newblock \emph{arXiv preprint arXiv:2508.09987}, 2025.

\bibitem[Zhang et~al.(2023{\natexlab{a}})Zhang, Mo, Chen, Sun, and Su]{Zhang2023MagicBrush}
Kai Zhang, Lingbo Mo, Wenhu Chen, Huan Sun, and Yu Su.
\newblock Magicbrush: A manually annotated dataset for instruction-guided image editing.
\newblock In \emph{Advances in Neural Information Processing Systems}, 2023{\natexlab{a}}.

\bibitem[Zhang et~al.(2023{\natexlab{b}})Zhang, Rao, and Agrawala]{zhang2023controlnet}
Lvmin Zhang, Anyi Rao, and Maneesh Agrawala.
\newblock Adding conditional control to text-to-image diffusion models.
\newblock In \emph{ICCV}, 2023{\natexlab{b}}.

\bibitem[Zhang et~al.(2025)Zhang, Xie, Lu, Yang, and Yang]{zhang2025icedit}
Zechuan Zhang, Ji Xie, Yu Lu, Zongxin Yang, and Yi Yang.
\newblock In-context edit: Enabling instructional image editing with in-context generation in large scale diffusion transformer.
\newblock \emph{arXiv preprint arXiv:2504.20690}, 2025.

\bibitem[Zhao et~al.(2024)Zhao, Ma, Chen, Si, Wu, An, Yu, Zhang, Li, and Chang]{zhao2024ultraedit}
Haozhe Zhao, Xiaojian~Shawn Ma, Liang Chen, Shuzheng Si, Rujie Wu, Kaikai An, Peiyu Yu, Minjia Zhang, Qing Li, and Baobao Chang.
\newblock Ultraedit: Instruction-based fine-grained image editing at scale.
\newblock \emph{Advances in Neural Information Processing Systems}, 37:\penalty0 3058--3093, 2024.

\bibitem[Zhou et~al.(2024)Zhou, Yu, Babu, Tirumala, Yasunaga, Shamis, Kahn, Ma, Zettlemoyer, and Levy]{zhou2024transfusion}
Chunting Zhou, Lili Yu, Arun Babu, Kushal Tirumala, Michihiro Yasunaga, Leonid Shamis, Jacob Kahn, Xuezhe Ma, Luke Zettlemoyer, and Omer Levy.
\newblock Transfusion: Predict the next token and diffuse images with one multi-modal model.
\newblock \emph{arXiv preprint arXiv:2408.11039}, 2024.

\end{thebibliography}
}


\end{document}